\documentclass[conference]{IEEEtran}
\IEEEoverridecommandlockouts

\usepackage{cite}
\usepackage{amsmath,amssymb,amsfonts}
\usepackage{algorithmic}
\usepackage{graphicx}
\usepackage{textcomp}
\usepackage{xcolor}
\usepackage{graphicx} 
\usepackage{amsmath} 
\usepackage{amsxtra} 
\usepackage{amssymb} 
\usepackage{amsthm} 
\usepackage{latexsym} 
\usepackage{setspace} 
\usepackage{nomencl} 
\usepackage[margin=1in]{geometry} 
\usepackage[titles]{tocloft} 
\usepackage{booktabs}
\usepackage{subcaption}
\usepackage{arydshln}

\def\BibTeX{{\rm B\kern-.05em{\sc i\kern-.025em b}\kern-.08em
    T\kern-.1667em\lower.7ex\hbox{E}\kern-.125emX}}
\begin{document}

\title{
Predictive Modeling and Explainable AI for Veterinary Safety Profiles, Residue Assessment, and Health Outcomes Using Real-World Data and Physicochemical Properties
\\
}




\author{Hossein Sholehrasa\textsuperscript{1,2}, 
Xuan Xu\textsuperscript{1,3}, 
Doina Caragea\textsuperscript{2}, 
Jim E. Riviere\textsuperscript{1},
Majid Jaberi-Douraki\textsuperscript{1,4,*}\thanks{*Corresponding Author: jaberi@k-state.edu}\\
\textsuperscript{1}1DATA Consortium and FARAD Program, Kansas State University, Olathe, KS, USA\\
\textsuperscript{2}Department of Computer Science, Kansas State University, Manhattan, KS, USA\\
\textsuperscript{3}Department of Statistics, Kansas State University, Manhattan, KS, USA\\
\textsuperscript{4}Department of Mathematics, Kansas State University, Olathe, KS, USA\\

}

\maketitle

\begin{abstract}
The safe use of pharmaceuticals in food-producing animals is critical for protecting animal welfare and human food safety. Adverse events (AEs) in veterinary medicine may reflect unexpected pharmacokinetic or toxicokinetic effects that result in violative residues in the food chain. This study introduces a predictive modeling framework to classify animal health outcomes as either Death or Recovered using the U.S. FDA’s OpenFDA Center for Veterinary Medicine database of $\sim$1.28 million reports (1987--2025 Q1). A comprehensive preprocessing pipeline linked relational tables and mapped AEs and drugs to standardized ontologies, including the Veterinary Dictionary for Drug Regulatory Activities (VeDDRA) and the WHO’s Anatomical Therapeutic Chemical (ATC) system for veterinary use. Data were standardized, missing values imputed, and high-cardinality variables reduced. Physicochemical drug properties were also integrated to capture relationships between chemical characteristics, distribution, and residue potential.
We evaluated multiple supervised models, including Random Forest, CatBoost, XGBoost, the transformer-based ExcelFormer, and in-context learning with large language models (Gemma 3-27B, Phi 3-12B). To address class imbalance, we applied undersampling, oversampling, and the Synthetic Minority Oversampling Technique (SMOTE) combined with Edited Nearest Neighbors (ENN), emphasizing recall for fatal outcomes. Ensemble methods achieved the strongest performance, with CatBoost, Voting, and Stacking classifiers each reaching precision, recall, and F1-scores of 0.95. Incorporating Average Uncertainty Margin (AUM)-based pseudo-labeling of uncertain cases further improved minority-class detection, particularly enhancing recall in ExcelFormer and XGBoost.
Model interpretability was provided through SHapley Additive exPlanations (SHAP), which identified biologically plausible predictors, including organ system disorders (e.g., bronchial, lung, and heart conditions), animal demographics, and drug physicochemical properties. These features were strongly linked to fatal outcomes and critical in distinguishing Death from Recovered cases. This work demonstrates that combining rigorous data engineering, advanced machine learning, and explainable AI enables accurate, interpretable prediction of veterinary safety outcomes, strengthening residue risk assessment in food-producing animals. The framework aligns with the mission of the Food Animal Residue Avoidance Databank (FARAD), supporting early detection of high-risk profiles and informing regulatory and clinical decision-making.

\end{abstract}


\section{Introduction}
The safe use of pharmaceuticals in animals is a fundamental component of public health protection, safeguarding animal welfare and, in the case of food-producing animals, ensuring human food safety. In food-producing animals, veterinary drugs, including antimicrobials, antiparasitics, anti-inflammatories, and biologics, are integral to maintaining food-producing animals' health and productivity \cite{mesfin2024veterinary}. Yet, their use carries the risk of leaving residues in edible tissues, milk, and other animal-derived products. Regulatory agencies such as the U.S. Food and Drug Administration (FDA) and European Medicines Agency (EMA) set maximum residue limits (MRLs) and withdrawal periods to ensure that these residues remain at or below levels considered safe for human consumption \cite{zad2023development}. Exceeding these limits can result in chronic dietary exposure, which may contribute to antimicrobial resistance (AMR) development, trigger allergic or toxic reactions in sensitive individuals, and, for certain compounds, pose potential carcinogenic or endocrine-disrupting risks \cite{mesfin2024veterinary}.
Beyond food safety, adverse events (AEs) in veterinary medicine are also a major concern for companion animals such as dogs and cats, where they can cause morbidity, mortality, and compromised welfare. In both food-producing animals and companion species, AEs may reflect unexpected pharmacokinetic (PK) or toxicokinetic behaviors that alter drug safety profiles. For example, impaired liver or kidney function can delay drug clearance, prolonging residue persistence and withdrawal times in food animals or increasing toxicity risks in pets \cite{khalifa2024veterinary}. Furthermore, cross-species differences in drug metabolism can complicate the extrapolation of safety data from one species to another, reinforcing the need for ongoing monitoring in real-world agricultural contexts \cite{margiotta2024cross}. In food animals, these complexities are directly tied to residue avoidance programs, where resources such as the Food Animal Residue Avoidance Databank (FARAD) play a central role in guiding withdrawal interval recommendations and supporting evidence-based decisions to minimize violative residues \cite{mercer2022mechanisms}.

Conventional pharmacovigilance (PV) largely depends on disproportionality analyses of spontaneous reporting system data, which are valuable for retrospective signal detection but insufficient for predicting emerging risks or managing complex, multidimensional datasets \cite{dijkstra2024discovery}. In contrast, machine learning (ML) and deep learning (DL) approaches can integrate diverse, high-dimensional datasets, including demographic, clinical, pharmacological, and AEs information, and automatically learn complex, non-linear relationships among variables. Unlike conventional statistical models that typically test predefined hypotheses and rely on simplified assumptions, ML/DL can uncover patterns without prior specification and adapt to new data for forecasting and real-time risk prediction. This shift enables proactive assessment and classification of animal health outcomes, with our predictive models for veterinary drugs advancing earlier detection of safety signals and exposure risks relevant to human consumers \cite{golmohammadi2025comprehensive}.

In this study, we develop an integrated predictive modeling framework for animal health outcome classification. Leveraging OpenFDA reports and PubChem's physicochemical properties, we apply supervised learning, semi-supervised Average Uncertainty Margin (AUM)-based pseudo-labeling, and explainable artificial intelligence (XAI) techniques to predict fatal versus non-fatal outcomes and identify the most influential predictors. Our work advances veterinary PV and human toxicology by demonstrating how predictive modeling can improve early detection of safety signals and provide insights relevant to protecting both animal and human health.

From a computational perspective, this research situates veterinary PV within the broader field of applied ML for complex biomedical datasets. The OpenFDA Center for Veterinary Medicine (CVM) dataset presents challenges common to large-scale, real-world health data, including heterogeneous formats, high-cardinality categorical variables, class imbalance, and incomplete or noisy entries. To address these issues, we developed a robust preprocessing pipeline in Section \ref{sec:preprocessing} that integrates relational datasets, maps data to standardized ontologies, incorporates the Veterinary Dictionary for Drug Regulatory Activities (VeDDRA), and applies hierarchical drug classification using the WHO Anatomical Therapeutic Chemical (ATC) system for veterinary use (ATCvet codes) \cite{sholehrasa2025leveraging}. We then employed a multi-model strategy spanning classical algorithms, ensemble methods (Random Forest, CatBoost, XGBoost), and transformer-based DL (ExcelFormer) to model non-linear interactions between demographic, pharmacological, and clinical features. Semi-supervised AUM-based pseudo-labeling enables the expansion of the training set with confidently predicted unlabeled cases, improving minority-class representation. At the same time, SHAP (SHapley Additive exPlanations) feature importance analyses ensure interpretability and clinical relevance. This integration of rigorous data engineering, advanced predictive modeling, and explainable AI provides a scalable and transparent computational framework for proactively identifying high-risk drug–event profiles in veterinary medicine.

\section{Related Work}
Recent advances in biomedical AI highlight the growing role of ML and XAI across domains such as disease detection, survival analysis, and diagnostic decision support. Shaon et al. \cite{shaon2024advanced} proposed a stacking ensemble for liver disease classification, demonstrating that ensemble methods can boost diagnostic accuracy, though their analysis was based on a small dataset with limited generalizability. Hashtarkhani et al. \cite{hashtarkhani2024explainable} developed an explainable pipeline for breast cancer survival prediction by integrating EMRs with socioeconomic and geographic data. Using a total of 10,172 data points, they provided insights into health disparities, yet their study was constrained to a single region and patient cohort. 

Computational PV research increasingly leverages large-scale AE databases to detect safety signals through advanced statistical analysis and ML. In the veterinary domain, Xu et al. \cite{xu2019making} applied ML to the OpenFDA veterinary AE dataset for dogs and cats, using selamectin and spinosad as case studies to show how similarity and network-based methods can reveal hidden drug–event patterns, distinguish severe from non-severe events, and validate labeling for improved post-market drug safety. More recently, Xu et al. \cite{xu2024silico} reviewed \textit{in-silico} strategies integrating disproportionality analyses, AI-driven clustering, pharmacogenomics, and advanced visualization, emphasizing systematic data curation and standardized terminology mapping as approaches directly applicable to this work. In parallel, most human-focused PV studies using FAERS and OpenFDA rely on disproportionality-based signal detection. Examples include \cite{pan2024disproportionality} for ticagrelor, \cite{yang2024adverse} for memantine + donepezil, and \cite{zhong2025multidimensional} for finasteride, all of which employed rigorous FAERS preprocessing and statistical algorithms such as the Proportional Reporting Ratio, Reporting Odds Ratio, and Bayesian Confidence Propagation Neural Network to characterize safety profiles. In contrast, predictive modeling has been less explored; Farnoush et al. \cite{farnoush2024prediction} integrated FAERS data with DrugBank \cite{knox2024drugbank} and PubChem \cite{kim2021pubchem} to train Random Forest and DL models for AE prediction using demographic and drug-chemical features.

Building on these methodological advances, our work applies predictive modeling principles to the OpenFDA CVM AE dataset, extending some of the PV research from retrospective human AE characterization to forward-looking veterinary safety profile and health outcome prediction. Unlike prior biomedical and PV studies that were limited by small, disease-specific datasets or narrow species/drug coverage, our work leverages the CVM OpenFDA database comprising more than {\it one million} reports with thousands of compounds across the nine most reported species, thereby addressing scalability and generalizability. To mitigate challenges of high-dimensional and noisy real-world data, our pipeline integrates relational tables, applies standardized ontologies (ATCvet, VeDDRA), and implements systematic handling of missing values and class imbalance. Moreover, where previous efforts primarily relied on retrospective disproportionality analyses or black-box predictive models, we extend the methodology through semi-supervised AUM-based pseudo-labeling to incorporate uncertain outcomes, improving minority-class representation and enabling more robust, forward-looking predictions. Finally, by embedding SHAP-based interpretability, our framework not only achieves high predictive performance but also provides transparent insights into model decision-making, helping to bridge the gap between accuracy and trustworthiness for regulatory and clinical applications.
%

\section{Data Preparation}

\subsection{Dataset Curation}
The primary dataset utilized in this research was obtained from the Animal and Veterinary Adverse Event Reporting System (AERS) maintained by the FDA, accessible through its public OpenFDA platform \cite{OpenFDA:DataDownloads}. This open-access repository provides one of the most comprehensive collections of animal reports in veterinary medicine. It serves as the foundation for an early-warning system to detect post-market safety issues with animal drugs. 
The dataset contains 1,279,530 reports for 94 species submitted between 1987 and 2025 Q1, sourced from veterinarians, animal owners, pharmaceutical companies, and academic institutions. The database is updated on a quarterly basis and provides structured information on animal demographics (species, breed, gender, age, weight), clinical outcomes (recovery, death, ongoing or unknown), drug details (active ingredients, brand name, dosage form, route of administration), and AEs (coded using standard medical ontologies for medical symptoms). Each report is uniquely identified by a unique\_aer\_id\_number to ensure consistent referencing across related tables.
The raw data, distributed in quarterly compressed JSON files following the OpenFDA API schema, was programmatically downloaded, parsed, and converted into four structured CSV files: the Main Table (metadata and demographics), AEs Table, Outcome Table (clinical outcomes and affected animals counts), and Drug Table (drug-specific information). In this analysis, CSV files were employed as the primary data format and subsequently imported into a PostgreSQL relational database to facilitate scalable querying and seamless integration during preprocessing. To efficiently load approximately one million rows, the pipeline first wrote the data into an in-memory buffer, which was then streamed in bulk via PostgreSQL’s COPY FROM STDIN command. This buffering strategy reduced disk I/O, allowed efficient memory-based transfer, and enabled the database engine to parallelize parsing and batch disk writes. As a result, the approach avoided the overhead of millions of individual INSERT statements and leveraged PostgreSQL’s write-ahead logging to ensure both scalability and reliability in large-scale data handling.

To complement the structured drug records, additional chemical descriptors for active ingredients were retrieved from the PubChem REST API \cite{kim2021pubchem}. The inclusion of such properties not only supports the mechanistic interpretation of AEs but also enables downstream modeling approaches that require quantitative representations of drug physicochemical profiles. Specifically, computed chemical and physical properties were extracted, including molecular weight, hydrogen bond acceptor count, and partition coefficient estimates (XLogP3). These descriptors were integrated into the Drug Table, thereby enriching the dataset with molecular-level features.

\subsection{Preprocessing Steps}
\label{sec:preprocessing}


\subsubsection{Dataset Integration and Merging}

The four tables generated during data curation were merged using the unique\_aer\_id\_number, yielding a unified and relationally consistent dataset. This integration ensured that each row captured the full context of a given report, linking species demographics, drug characteristics, reported AEs, and final clinical outcomes. Because a single report may contain multiple drugs and AEs, categorical values (e.g., AEs, active ingredient names, routes of administration) were concatenated using the `\textbackslash{}` separator to preserve all entries in a compact format. For numerical descriptors, including drug-related chemical and physical properties obtained from PubChem, values were aggregated via summation across all active ingredients in the report, as this reflects the total molecular exposure of multi-ingredient formulations. Summation was preferred over mean or maximum, which would obscure additive effects and fail to capture the increased chemical load of combination therapies.

\subsubsection{Terminology Standardization}
The dataset incorporates internationally recognized veterinary coding systems to ensure consistency, comparability, and regulatory compliance. AEs were standardized using the VeDDRA hierarchy, which organizes descriptors from Lowest Level Terms (LLT) to Preferred Terms (PT), then into High-Level Terms (HLT) based on anatomical or functional similarity, and finally into broad System Organ Classes (SOC) covering major physiological systems \cite{bousquet2005appraisal}. Drug information, in turn, was classified using the ATCvet system, developed by the WHO Collaborating Centre (WHOCC) for drug statistics methodology. ATCvet assigns drugs to five hierarchical levels: Anatomical Main Group, Therapeutic Subgroup, Pharmacological Subgroup, Chemical Subgroup, and Chemical Substance, ensuring consistent representation across compounds \cite{dahlin2001atcvet}.

Building on these frameworks, we applied a standardized terminology mapping process to harmonize AEs and drug classes across reports. For AEs, PTs, and LLTs were mapped to HLTs to reduce granularity, consolidate synonymous or highly specific terms. For drugs, ATCvet codes were systematically mapped to the Chemical Subgroup level using automated web scraping from the WHOCC database \cite{WHOCC_ATCvet}. This approach ensured structured, regulatory-compliant data suitable for downstream modeling, while enabling pattern discovery across broader chemical categories without being constrained by overly specific substance identifiers.

\subsubsection{Data Cleaning and Normalization}
The dataset comprises reports, with every animal described by 92 structured features. The features vary in cardinality, from low-dimensional attributes such as Animal Gender (4 categories) to high-cardinality fields such as active ingredient names ($>$18,500 categories) and AEs ($>$3,000 categories).
To ensure consistency across records, several columns experienced systematic normalization and cleaning during preprocessing. For example, animal age values were converted to a consistent unit of years by transforming entries reported initially in days, weeks, or months. Animal weight was standardized to kilograms. In most reports, only the minimum age and weight fields were populated, while the corresponding maximum fields were frequently left empty. Missing values in animal age and weight were imputed using species-specific means, ensuring that each imputed value reflected the typical demographic profile of the corresponding species.
Beyond numerical data cleaning, several categorical features in the dataset contained missing or noisy values, which were addressed through targeted species-species cleaning strategies. Missing entries in columns like gender, dosage form, and route of administration were filled using the most frequently occurring value within each species. In addition, the reports with the ``Euthanized" category were removed due to insufficient information for accurate classification. The ``Recovered with Sequela" and ``Recovered/Normal" outcome categories were combined to avoid overlap between recovery types.

\subsubsection{Feature Engineering and Representation}
To improve the interpretability of the model and minimize noise, a manual feature filtering technique was applied. The Year feature was excluded to avoid introducing temporal bias, and reports with AE of ``Lack of Efficacy" were removed since they reflect treatment failure rather than a physiological response, which could potentially leak outcome labels.
All categorical features were encoded numerically (label encoding) to represent their categories as integer values while maintaining compatibility with the ML and DL models. Furthermore, Pearson correlation analysis was performed on the numeric features, and pairs of variables with high correlation coefficients were pruned to reduce redundancy and multicollinearity. In such cases, only one representative feature was retained. For example, highly correlated PubChem descriptors such as exact mass and molecular weight were compared, with molecular weight preserved as the representative variable.

\subsection{Dataset Partitioning}
After cleaning and feature engineering, the dataset contained approximately 690,000 reports, and it was then divided into two main groups based on outcome labels. The first group contained reports with definitive outcomes, specifically Death and Recovered cases ($\sim 280,000$), while the second group consisted of reports with uncertain outcomes categorized as Ongoing or Unknown ($\sim 410,000$). Using a standard stratified sampling strategy, we divided the labeled dataset into three subsets: 80\% for training, 10\% for validation, and 10\% for testing \cite{geron2022hands}. The training set was employed to build the initial model, the validation set was used for hyperparameter tuning, and the test set was held out to assess the final model’s performance.

\subsection{Handling Class Imbalance}

The dataset showed a class imbalance, with Recovered cases outnumbering Death cases (85\% vs 15\%). To address this, we independently evaluated several sampling strategies: Random Oversampling, Random Undersampling, and hybrid approaches combining oversampling with noise reduction. Specifically, we applied the Synthetic Minority Oversampling Technique with Edited Nearest Neighbors (SMOTE+ENN) \cite{yang2022hybrid}, where synthetic minority samples are generated and noisy majority samples removed. Each method was tested separately to assess its impact on performance, enabling selection of the most effective approach for improving generalization and enhancing minority-class recall.

\section{Methodology}


\subsection{Supervised Learning Models}
\subsubsection{Traditional ML \& Ensembles}
The ML component of this study incorporated a diverse set of classification algorithms to classify reports into either Death or Recovered.. These included traditional linear models such as Logistic Regression, and tree-based methods including Decision Tree, Random Forest, AdaBoost, CatBoost, and XGBoost. Additionally, the K-Nearest Neighbors (KNN) algorithm was employed to leverage instance-based learning for outcome prediction.

We also used ensemble learning techniques to strengthen the robustness and generalizability of our predictive models by combining multiple base learners to reduce bias and leverage complementary strengths. Specifically, we applied a soft Voting Classifier \cite{geron2022hands} with Catboost, Random Forest, and Xgboost, and a Stacking Classifier \cite{wolpert1992stacked} using the same base models with a logistic regression meta-learner.

\subsubsection{DL \& LLM Models}
The study also incorporated a Multilayer Perceptron (MLP) model, implemented as a feedforward neural network with fully connected layers and trained with the Adam optimizer, and the ExcelFormer model \cite{chen2023excelformer}, a transformer-based architecture designed for tabular data. Both models were fine-tuned on training data, hyperparameter-tuned using the validation data, and finally evaluated on the test set to benchmark their performance against that of the classical ML and ensemble models.

To complement ML and DL approaches, we further explored the use of LLMs, such as Gemma 3-27B and Phi 3-12B, for predictive inference on our task. The motivation for this step was to leverage the in-context reasoning ability of LLMs to capture complex relationships between heterogeneous clinical and chemical features that may not be fully exploited by conventional tabular models. We performed both 5-shot and zero-shot in-context learning experiments by converting the structured dataset into a text-based format suitable for LLM input. Specifically, each row was represented as a sequence of column–value pairs, concatenated using the `\textbar` delimiter (e.g., \textit{age: 5 \textbar\ gender: female \textbar\ species: Dog \textbar\ molecular weight: 312.4}). These textual representations were then provided as part of prompts to the LLM for row-level inference. This design allowed the model to directly process structured reports in natural language form, offering an alternative perspective for outcome prediction and a means to evaluate the generalization capacity of LLMs in low-resource and heterogeneous biomedical settings.


\subsection{Semi-supervised Learning Approach}

To incorporate reports with Unknown or Ongoing outcomes into the modeling process, we adopted a pseudo-labeling strategy \cite{lee2013pseudo}. In this approach, a baseline model was first trained on labeled data (Death and Recovered) and then used to assign labels to the unlabeled cases. The resulting pseudo-labels were combined with the original labeled set for retraining, allowing the model to iteratively improve its performance by leveraging a larger training set.

To improve the reliability of pseudo-label selection, we applied the AUM metric \cite{pleiss2020identifying}. AUM evaluates the stability of predictions across training epochs by averaging the difference between the top two predicted class probabilities:
\[
\text{AUM} = \frac{1}{T} \sum_{t=1}^{T} \left( P_{\text{top}}^t - P_{\text{second}}^t \right).
\]
Pseudo-labeled samples were ranked according to each metric, and only the top fraction (20--80\%, depending on validation results) was retained for retraining. 
These complementary strategies ensure that pseudo-labels incorporated into the SSL pipeline are supported by strong evidence of confidence, thereby reducing the propagation of noise.

\begin{figure}[htbp]
    \centering
    \includegraphics[width=\columnwidth]{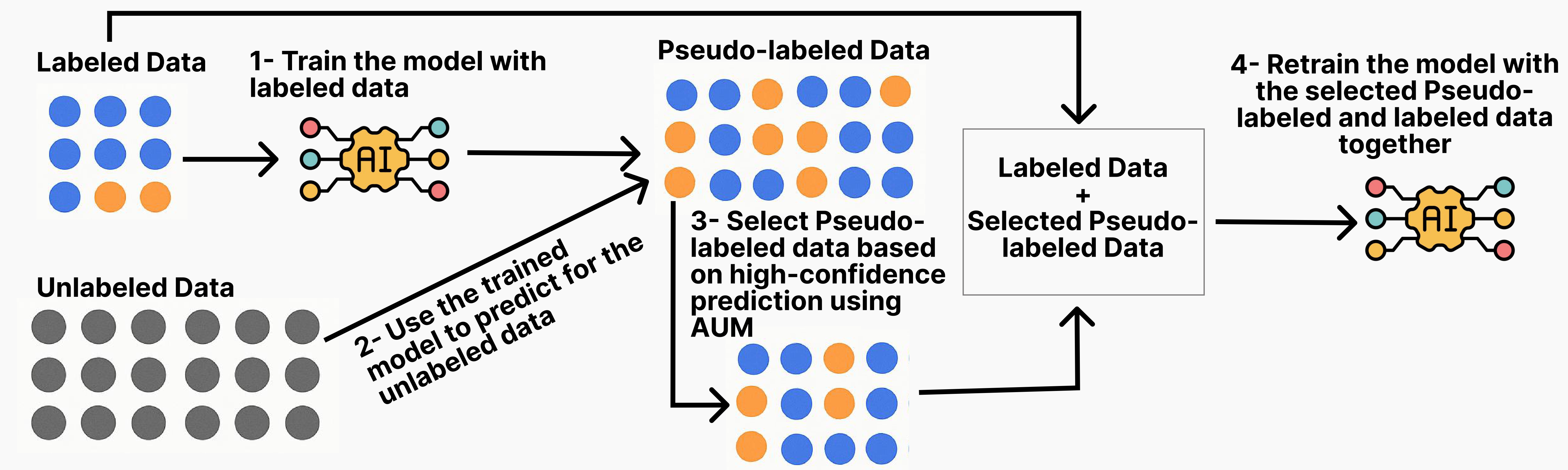}
    \caption{Overview of the SSL pipeline, illustrated with an AUM-based pseudo-labeling workflow. The model is trained on labeled data, assigns pseudo-labels to unlabeled cases, and high-confidence predictions are merged with the labeled set for retraining.}
    \vspace{-1.5em}
    \label{fig:pseudo_labeling}
\end{figure}

The final SSL pipeline integrated pseudo-labeling with AUM-based high-confidence filtering. 
As illustrated in Figure~\ref{fig:pseudo_labeling}, the model was first trained on labeled data, then used to generate pseudo-labels for unlabeled cases. Pseudo-labeled samples were ranked by each confidence metric (AUM), and only the high-confidence subset was merged with the labeled dataset for retraining. This integration expands the training set while reducing the risk of noisy labels, with different metrics providing alternative views of confidence.

\subsection{Evaluation metrics}

All models were evaluated using four weighted metrics: accuracy, F1-score, precision, and recall for both the Death and Recovered classes. These metrics provided a balanced assessment of overall performance, class-specific sensitivity, and the ability to handle class imbalance effectively.

\subsection{Model explainability}
To ensure transparency and interpretability in veterinary PV, we applied SHAP values to explain model predictions at the local instance-level \cite{lundberg2018consistent}. In this study, SHAP was applied to the test dataset to explain individual predictions, validate the model, and diagnose errors through visualizations. Furthermore, SHAP was applied to both ML classifiers and the transformer-based ExcelFormer. By quantifying how individual features (e.g., clinical variables or tabular embeddings) influenced the likelihood of Death versus Recovered outcomes, SHAP provided local explanations that revealed the drivers of predictions.
SHAP-based interpretability enhanced model transparency and yielded clinically relevant insights into veterinary drug safety by identifying the key factors associated with Death and Recovered outcomes.

\section{Results \& Discussion}

\subsection{Model Performance}
Table \ref{tab:all-results} summarizes the performance of all supervised and semi-supervised learning models under different sampling strategies: no sampling, undersampling, oversampling, and SMOTE+ENN. Performance was evaluated using five metrics: average weighted F1-score(F1), precision (P), recall (R), Death recall (DR), and Recovered recall (RR). Overall, ensemble-based approaches such as Random Forest, CatBoost, XGBoost, Voting Classifier, and Stacking Classifier consistently outperformed other models across all sampling strategies. These methods achieved the highest F1-scores (0.94 to 0.95), with a particularly strong balance between precision and recall. These models also demonstrated strong class-specific performance, achieving DR values above 0.77 and RR values above 0.90, indicating robustness in handling both minority and majority outcomes. These results underscore the capacity of these models to capture complex, non-linear interactions across heterogeneous clinical, physicochemical properties, and demographic features. The application of sampling strategies produced nuanced effects. Undersampling yielded balanced improvements in the minority class (Death) recall across several models. For example, Decision Trees improved from DR = 0.73 (no sampling) to 0.87 with undersampling, indicating the utility of rebalancing the training data. Similarly, ExcelFormer, the transformer-based architecture, demonstrated a notable boost in Death Recall under SMOTE+ENN and oversampling (DR = 0.86) compared to no sampling (DR = 0.64), though at the cost of reduced Death Precision. This highlights the trade-off between sensitivity to fatal outcomes and specificity in predictions. Undersampling, on the other hand, tended to lower overall F1-scores compared to no sampling or synthetic approaches.

Neural approaches such as the MLP achieved moderate performance (with F1 between 0.85 and 0.88 for different settings) but lagged behind ensemble methods, suggesting that shallow feedforward architectures are less effective for structured veterinary AE data. In contrast, ExcelFormer consistently delivered more competitive results (with F1 between 0.86 and 0.91), particularly under SMOTE+ENN, where it achieved its strongest balance of DR = 0.86 with F1 = 0.89, reinforcing the promise of transformer-based architectures for tabular biomedical tasks. However, DL approaches, their death recall values (from 0.64 to 0.86 for Excelformer and 0.46 to 0.72 in MLP) were less consistent across sampling techniques, suggesting some sensitivity to class imbalance handling.

\setlength{\tabcolsep}{2pt}
\begin{table*}[ht]
\centering
\tiny
\caption{Supervised and semi-supervised classification results under different resampling strategies (no sampling, undersampling, oversampling, and SMOTE+ENN). Reported metrics include weighted F1 average (F1), weighted precision (P), weighted recall (R), Death Recall (DR), and Recovered Recall (RR). Semi-supervised learning applied top 30\% AUM-based pseudo-labeling.}

\label{tab:all-results}
\resizebox{\textwidth}{!}{%
\begin{tabular}{lcccccccccccccccccccc}
\toprule
 & \multicolumn{20}{c}{\textbf{Supervised Learning}} \\
\cmidrule(lr){2-21}
 & \multicolumn{5}{c}{\textbf{No sampling}}
 & \multicolumn{5}{c}{\textbf{Undersampling}} 
 & \multicolumn{5}{c}{\textbf{Oversampling}} 
 & \multicolumn{5}{c}{\textbf{SMOTE + ENN}} \\
\cmidrule(lr){2-6} \cmidrule(lr){7-11} \cmidrule(lr){12-16} \cmidrule(lr){17-21}
\textbf{Model} & \textbf{F1} & \textbf{P} & \textbf{R} & \textbf{DR} & \textbf{RR}
               & \textbf{F1} & \textbf{P} & \textbf{R} & \textbf{DR} & \textbf{RR}
               & \textbf{F1} & \textbf{P} & \textbf{R} & \textbf{DR} & \textbf{RR}
               & \textbf{F1} & \textbf{P} & \textbf{R} & \textbf{DR} & \textbf{RR} \\
\midrule
Logistic Regression     & 0.85 & 0.85 & 0.84 & 0.51 & 0.90 & 0.84 & 0.85 & 0.83 & 0.56 & 0.88 & 0.85 & 0.85 & 0.84 & 0.52 & 0.90 & 0.84 & 0.85 & 0.83 & 0.57 & 0.88 \\
Decision Tree           & 0.92 & 0.92 & 0.92 & 0.73 & 0.95 & 0.88 & 0.91 & 0.86 & 0.87 & 0.86 & 0.92 & 0.92 & 0.92 & 0.72 & 0.95 & 0.91 & 0.92 & 0.90 & 0.85 & 0.91 \\
K-Nearest Neighbors     & 0.91 & 0.91 & 0.90 & 0.77 & 0.93 & 0.89 & 0.90 & 0.89 & 0.75 & 0.91 & 0.90 & 0.91 & 0.90 & 0.77 & 0.92 & 0.90 & 0.91 & 0.89 & 0.80 & 0.91 \\
Random Forest           & 0.94 & 0.94 & 0.94 & 0.81 & 0.96 & 0.94 & 0.94 & 0.94 & 0.77 & 0.97 & 0.94 & 0.94 & 0.94 & 0.80 & 0.97 & 0.93 & 0.93 & 0.93 & 0.76 & 0.96 \\
AdaBoost                & 0.90 & 0.90 & 0.91 & 0.58 & 0.97 & 0.90 & 0.90 & 0.91 & 0.53 & 0.98 & 0.90 & 0.90 & 0.91 & 0.53 & 0.97 & 0.90 & 0.90 & 0.90 & 0.62 & 0.95 \\
CatBoost                & 0.95 & 0.95 & 0.95 & 0.83 & 0.96 & 0.94 & 0.94 & 0.94 & 0.79 & 0.97 & 0.95 & 0.95 & 0.95 & 0.81 & 0.97 & 0.94 & 0.94 & 0.94 & 0.80 & 0.96 \\
XGBoost                 & 0.95 & 0.94 & 0.95 & 0.77 & 0.98 & 0.94 & 0.94 & 0.94 & 0.78 & 0.97 & 0.94 & 0.94 & 0.94 & 0.80 & 0.97 & 0.94 & 0.94 & 0.94 & 0.77 & 0.97 \\
\hline
Voting Classifier       & 0.95 & 0.95 & 0.95 & 0.82 & 0.97 & 0.94 & 0.94 & 0.95 & 0.80 & 0.97 & 0.95 & 0.95 & 0.95 & 0.81 & 0.97 & 0.94 & 0.94 & 0.94 & 0.79 & 0.97 \\
Stacking Classifier     & 0.95 & 0.95 & 0.95 & 0.82 & 0.97 & 0.94 & 0.94 & 0.95 & 0.80 & 0.97 & 0.94 & 0.94 & 0.94 & 0.76 & 0.97 & 0.94 & 0.94 & 0.94 & 0.78 & 0.96 \\
\hline
Multilayer Perceptron   & 0.87 & 0.87 & 0.87 & 0.56 & 0.92 & 0.85 & 0.85 & 0.85 & 0.46 & 0.92 & 0.88 & 0.89 & 0.88 & 0.65 & 0.92 & 0.87 & 0.89 & 0.87 & 0.72 & 0.89 \\
ExcelFormer             & 0.91 & 0.91 & 0.92 & 0.64 & 0.96 & 0.86 & 0.88 & 0.85 & 0.72 & 0.87 & 0.88 & 0.91 & 0.87 & 0.86 & 0.87 & 0.89 & 0.91 & 0.88 & 0.86 & 0.88 \\
\hline
Gemma 3-27B (0-shot) & 0.82 & 0.82 & 0.82 & 0.37 & 0.90 & - & - & - & - & - & - & - & - & - & - & - & - & - & - & - \\
Gemma 3-27B (5-shot) & 0.81 & 0.86 & 0.78 & 0.73 & 0.79 & - & - & - & - & - & - & - & - & - & - & - & - & - & - & - \\
Phi 3-12B (0-shot)   & 0.81 & 0.82 & 0.86 & 0.09 & 0.99 & - & - & - & - & - & - & - & - & - & - & - & - & - & - & - \\
Phi 3-12B (5-shot)   & 0.80 & 0.79 & 0.85 & 0.07 & 0.98 & - & - & - & - & - & - & - & - & - & - & - & - & - & - & - \\
\midrule
\multicolumn{21}{c}{\textbf{Semi-supervised (AUM-based pseudo-labeling)}} \\
\midrule
Decision Tree           & 0.91 & 0.91 & 0.91 & 0.73 & 0.94 & 0.89 & 0.91 & 0.88 & 0.83 & 0.89 & 0.91 & 0.91 & 0.91 & 0.71 & 0.95 & 0.91 & 0.92 & 0.91 & 0.83 & 0.92 \\
Random Forest           & 0.94 & 0.94 & 0.94 & 0.79 & 0.97 & 0.94 & 0.94 & 0.94 & 0.76 & 0.97 & 0.94 & 0.94 & 0.94 & 0.77 & 0.97 & 0.93 & 0.93 & 0.93 & 0.77 & 0.96 \\
CatBoost                & 0.95 & 0.95 & 0.95 & 0.80 & 0.97 & 0.94 & 0.94 & 0.94 & 0.78 & 0.97 & 0.95 & 0.95 & 0.95 & 0.79 & 0.98 & 0.94 & 0.94 & 0.94 & 0.78 & 0.97 \\
XGBoost                 & 0.94 & 0.94 & 0.94 & 0.80 & 0.97 & 0.94 & 0.94 & 0.94 & 0.79 & 0.97 & 0.94 & 0.94 & 0.94 & 0.79 & 0.97 & 0.93 & 0.93 & 0.93 & 0.78 & 0.96 \\
\hline
Voting Classifier       & 0.95 & 0.95 & 0.95 & 0.80 & 0.97 & 0.94 & 0.94 & 0.94 & 0.80 & 0.97 & 0.95 & 0.95 & 0.95 & 0.80 & 0.98 & 0.94 & 0.94 & 0.94 & 0.79 & 0.97 \\
Stacking Classifier     & 0.95 & 0.95 & 0.95 & 0.81 & 0.97 & 0.94 & 0.94 & 0.94 & 0.76 & 0.98 & 0.94 & 0.94 & 0.94 & 0.74 & 0.97 & 0.94 & 0.94 & 0.94 & 0.79 & 0.96 \\
\hline
ExcelFormer             & 0.92 & 0.91 & 0.92 & 0.63 & 0.97 & 0.86 & 0.89 & 0.85 & 0.82 & 0.85 & 0.91 & 0.91 & 0.90 & 0.80 & 0.92 & 0.91 & 0.92 & 0.91 & 0.80 & 0.93 \\
\bottomrule
\end{tabular}%
}
\vspace{-2em}
\end{table*}

\setlength{\tabcolsep}{6pt}

LLMs evaluated in zero-shot and few-shot settings underperformed relative to fine-tuned tabular models. For instance, Gemma 3-27B (5-shot) achieved an F1 of 0.81 with R = 0.78 and DR = 0.73, whereas Phi 3-12B (5-shot) reached F1 = 0.80 with R = 0.85 but a much lower DR = 0.07. These findings underscore the limitations of applying general-purpose LLMs to structured AE datasets without fine-tuning or domain-specific adaptation. Notably, for Gemma 3-27B, the inclusion of examples substantially improved DR (0.73 vs. 0.37 in zero-shot), while maintaining a strong F1 of 0.81.

Integrating unlabeled cases through AUM-based pseudo-labeling improved model robustness, particularly in recall. Under this strategy, XGBoost achieved F1 = 0.94, with an improvement in recall from 0.77 to 0.80 in the no sampling setting. Notably, ExcelFormer benefitted substantially from pseudo-labeling under SMOTE+ENN, improving to F1 = 0.91 with R = 0.91, up from R = 0.88 without augmentation. These findings demonstrate that leveraging uncertain reports via confidence-based selection can enhance minority-class representation across some models, particularly for deep models that benefit from larger, more diverse training sets. On the other hand, Catboost, Stacking, and Voting classifiers have achieved the strongest balance (F1=0.95, P=0.95, R=0.95), which semi-supervised learning maintained. For other models, however, slight declines in F1 were observed, likely due to label noise introduced by pseudo-labeling, despite being trained on a larger pool of reports.

In sum, ensemble methods such as CatBoost, XGBoost, and Stacking delivered the most reliable performance (F1 = 0.94-0.95) with strong balance across metrics. ExcelFormer was competitive and benefited from semi-supervised augmentation, improving recall under SMOTE+ENN, which highlights the potential of transformer-based models when trained with larger, diverse data. From a clinical perspective, enhancing Death recall is most critical, as failing to detect fatal outcomes carries higher regulatory and welfare risks than misclassifying recovery cases. Undersampling and semi-supervised pseudo-labeling both enhanced minority-class detection, with the latter providing a scalable strategy to incorporate large volumes of uncertain cases without significant loss in overall accuracy over some models. The limited performance of general-purpose LLMs underscores the need for domain-specific adaptation or fine-tuning when applying text-based reasoning to structured AE data. While LLMs hold promise for multi-modal integration (e.g., combining structured and unstructured case narratives), their raw application to tabular inputs remains suboptimal relative to purpose-built ML/DL methods.

\subsection{Model Interpretability}

\begin{figure*}[htbp]
    \vspace{-3ex}
    \centering
    \subfloat[SHAP summary plot for companion animals]{%
        \includegraphics[width=0.32\textwidth]{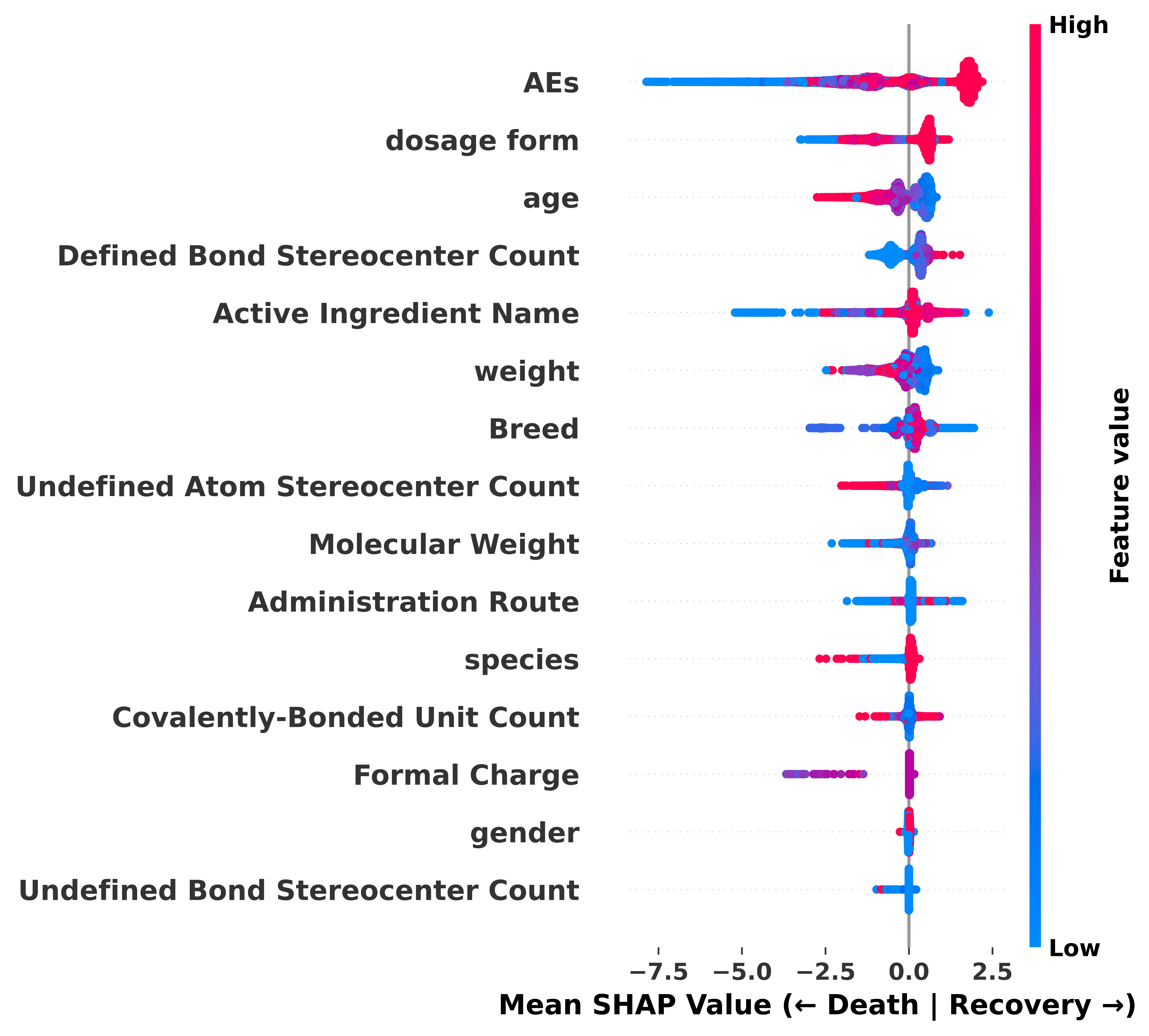}
    }\hfill
    \subfloat[SHAP summary plot for livestock]{%
        \includegraphics[width=0.32\textwidth]{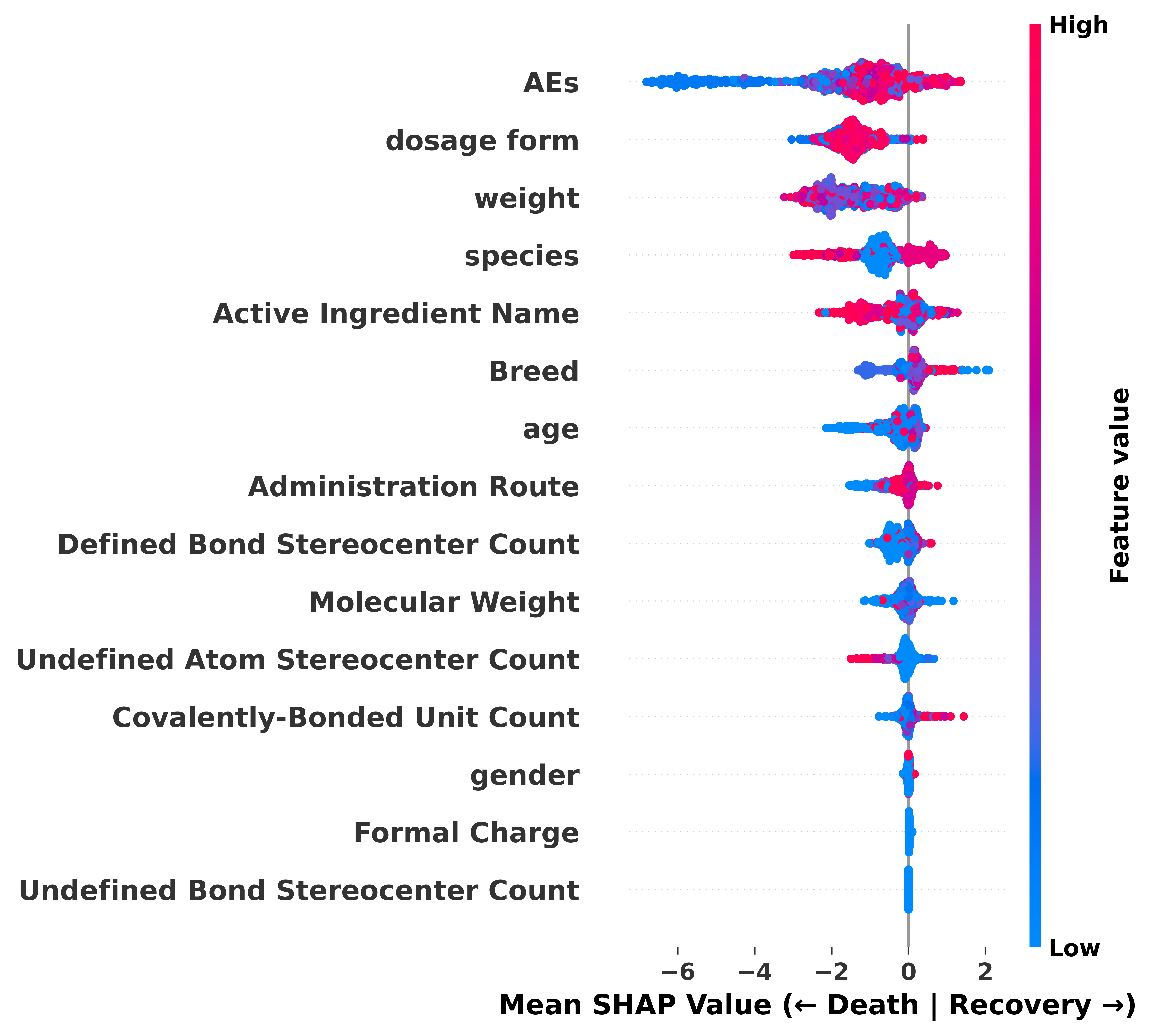}
    }\hfill
    \subfloat[SHAP summary plot for poultry]{%
        \includegraphics[width=0.32\textwidth]{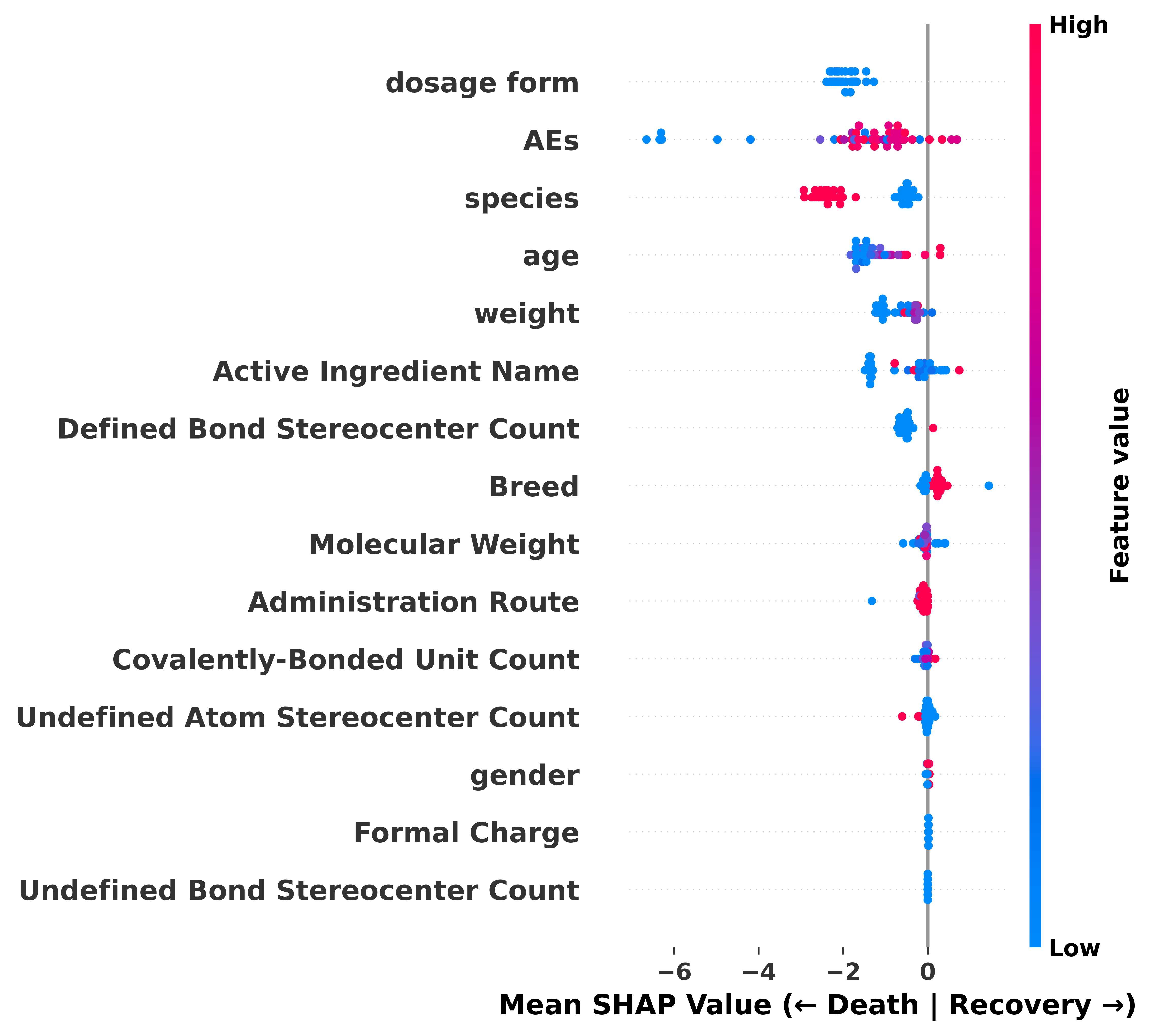}
    }\\[1ex]

    \subfloat[Top \& bottom SHAP mean values for AE terms (companion)]{%
        \includegraphics[width=0.32\textwidth]{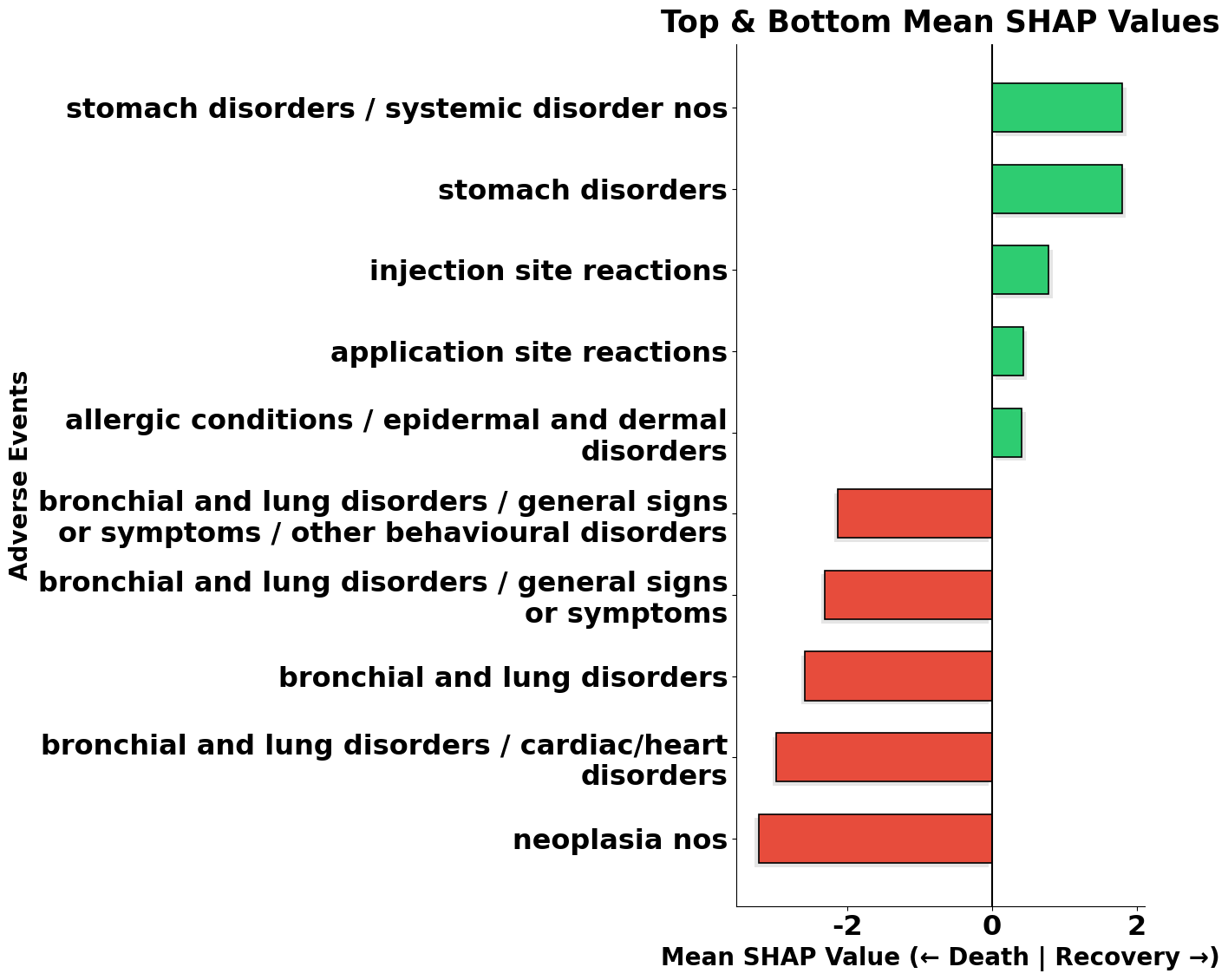}
    }\hfill
    \subfloat[Top \& bottom SHAP mean values for AE terms (livestock)]{%
        \includegraphics[width=0.32\textwidth]{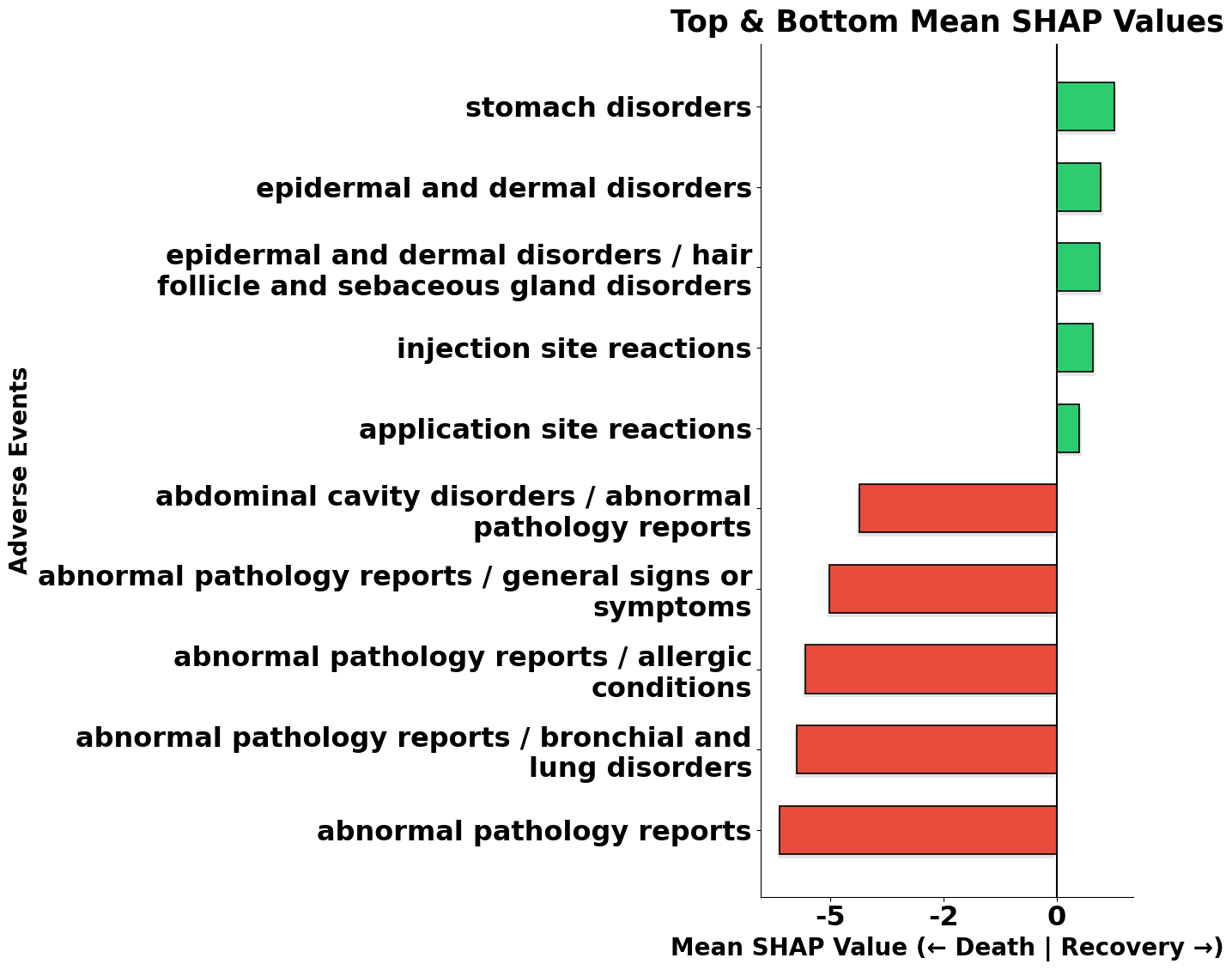}
    }\hfill
    \subfloat[Top \& bottom SHAP mean values for AE terms (poultry)]{%
        \includegraphics[width=0.32\textwidth]{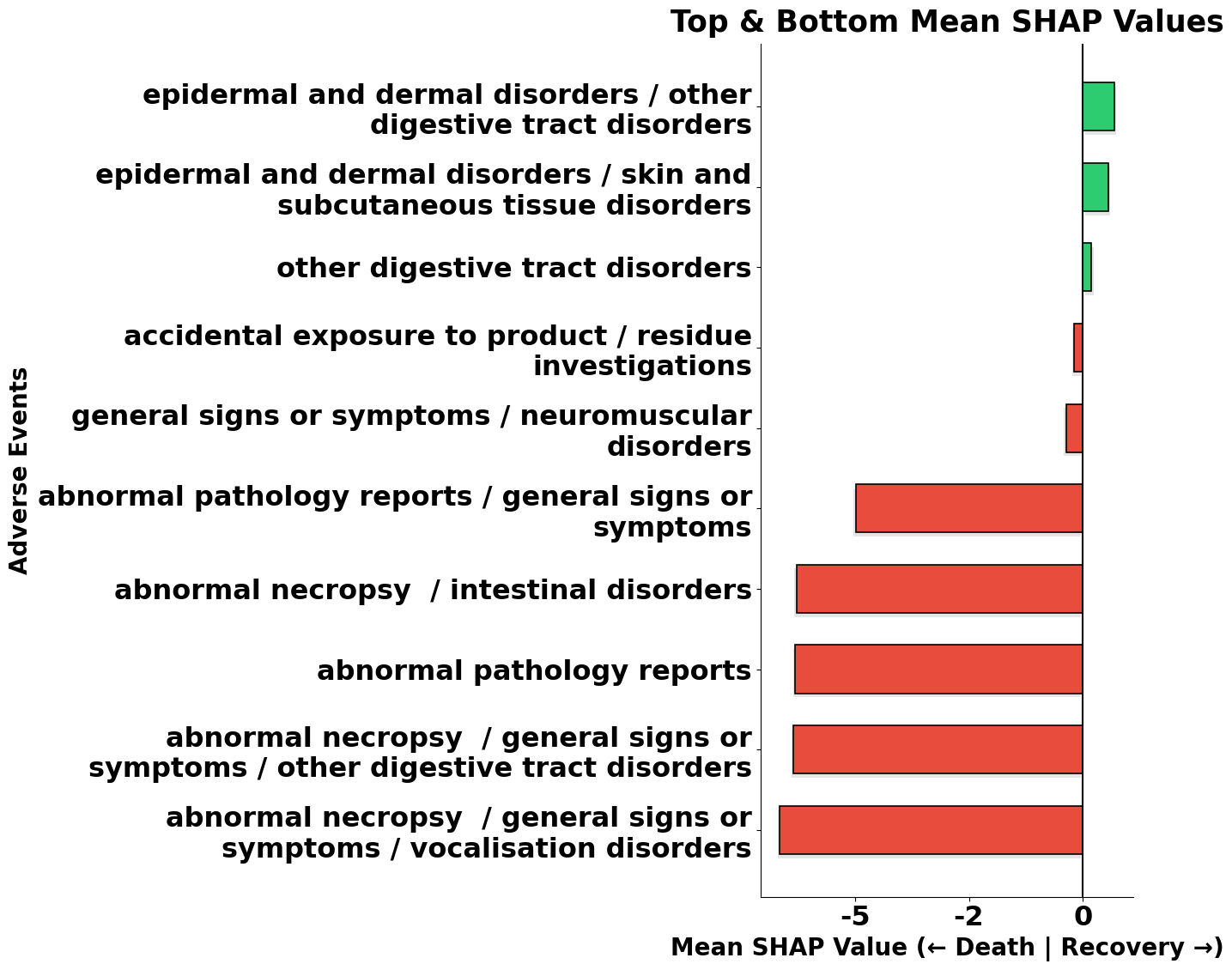}
    }\\[1ex]

    \subfloat[Top \& bottom SHAP mean values for active ingredients (companion)]{%
        \includegraphics[width=0.32\textwidth]{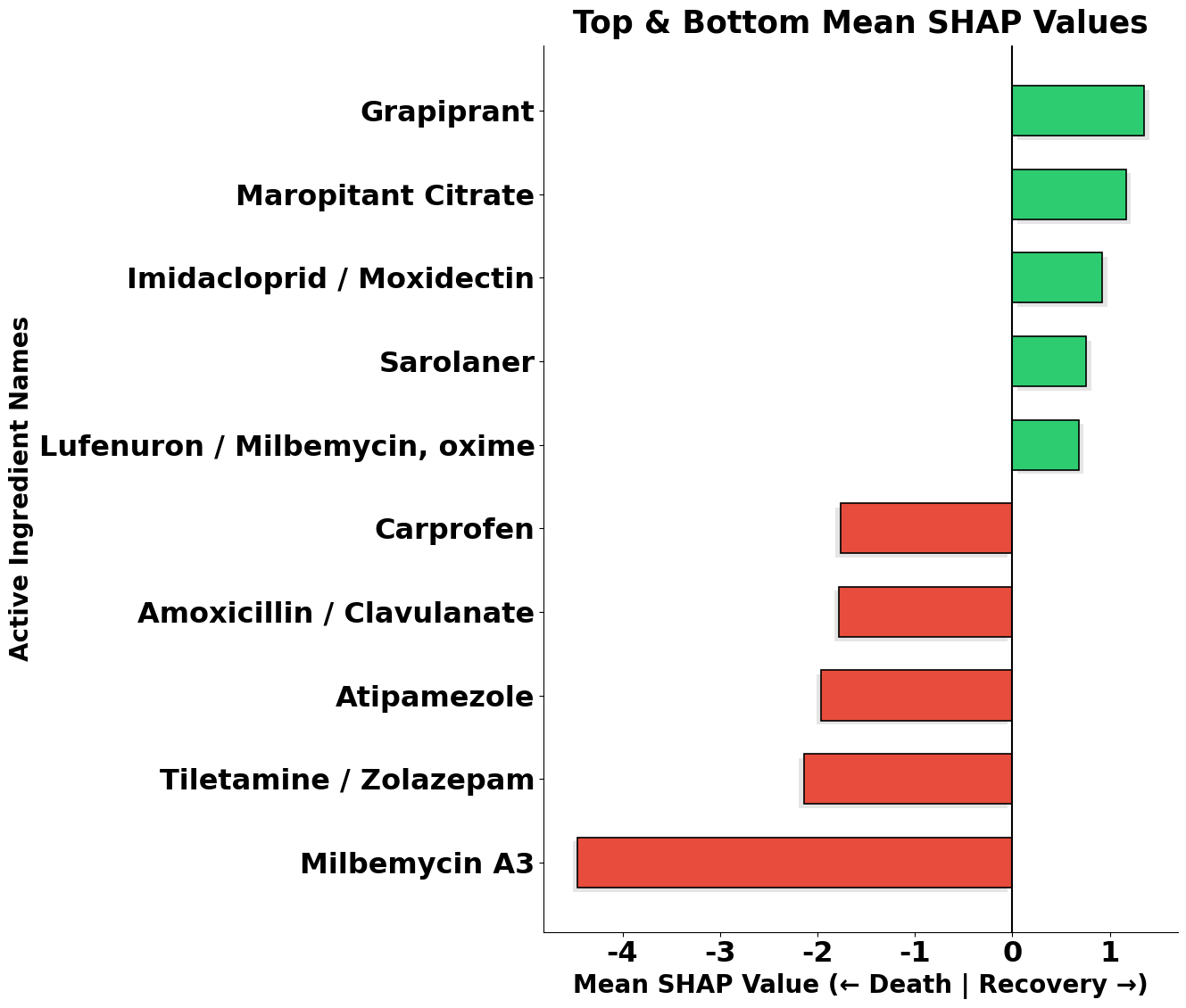}
    }\hfill
    \subfloat[Top \& bottom SHAP mean values for active ingredients (livestock)]{%
        \includegraphics[width=0.32\textwidth]{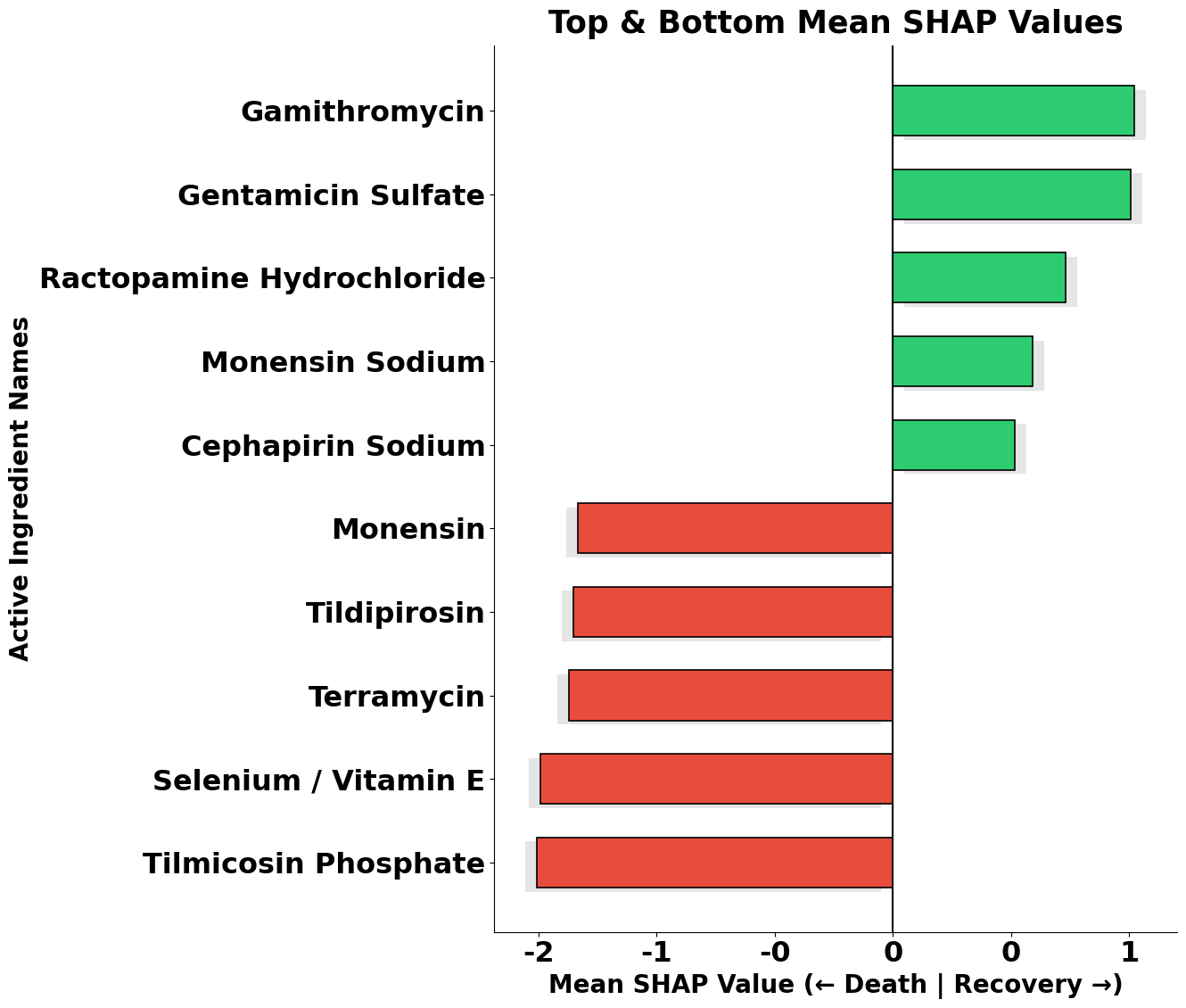}
    }\hfill
    \subfloat[Top \& bottom SHAP mean values for active ingredients (poultry)]{%
        \includegraphics[width=0.32\textwidth]{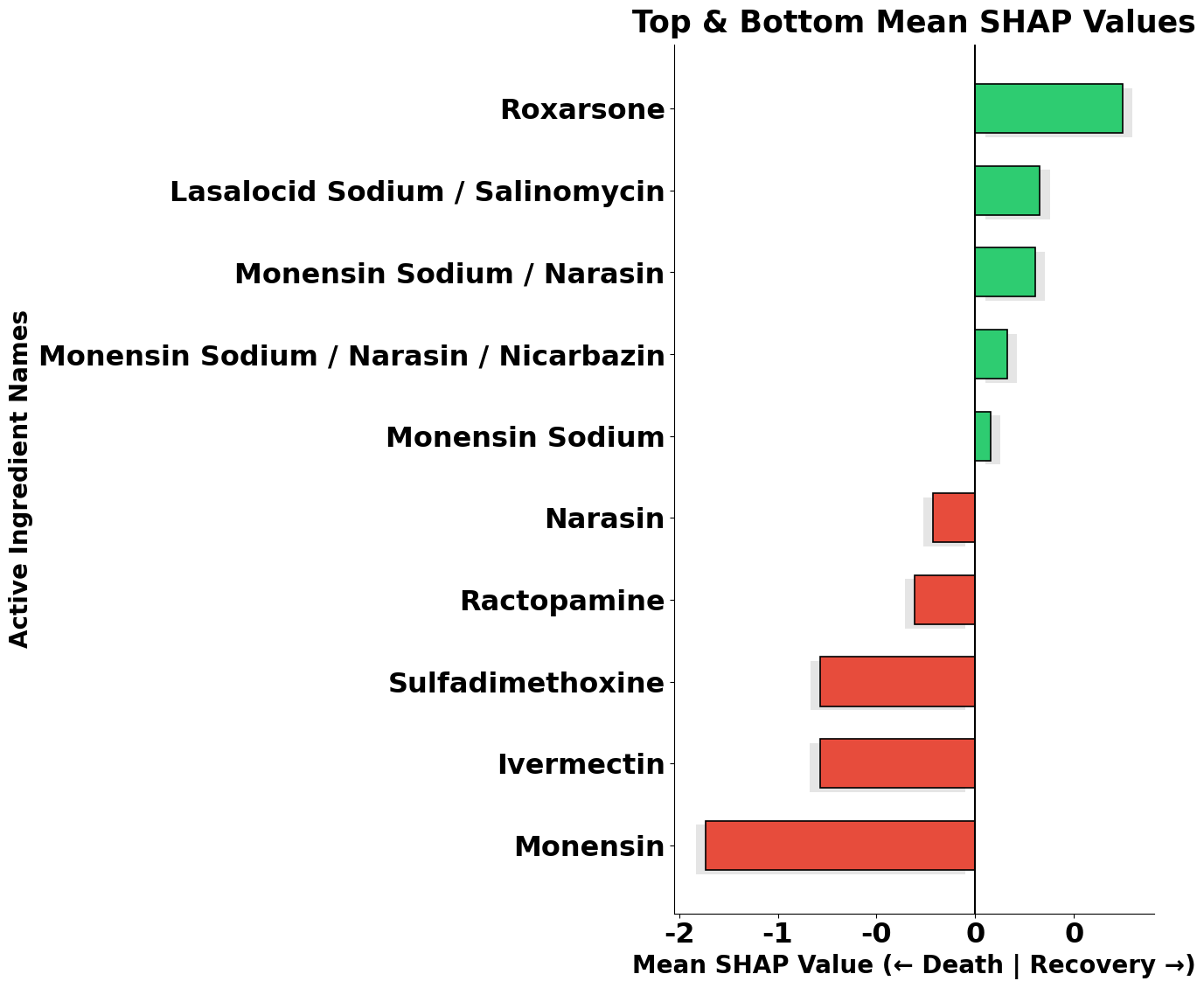}
    }

    \caption[Comprehensive SHAP visualizations across animal groups, drugs, and AEs.]
    {Comprehensive SHAP visualizations: (a–c) summary plots of the top 15 features for companion animals, livestock, and poultry; (d–f) top and bottom SHAP mean values for AE terms; (g–i) top and bottom SHAP mean values for active ingredients across animal groups.}
    \vspace{-1em}
    \label{fig:shap_summary_groups}
\end{figure*}

Figure~\ref{fig:shap_summary_groups} (a-c) presents SHAP summary plots for the Catboost model across three animal groups: companion animals, livestock, and poultry. Each point corresponds to an instance in the dataset, positioned according to its SHAP value, with negative values (left) indicating a higher likelihood of predicting Death and positive values (right) indicating a higher likelihood of predicting Recovered. Color encodes the feature value, ranging from low (blue) to high (red). Across all groups, AEs consistently emerged as the most influential predictors, spanning a broad range of SHAP values and driving both Death and Recovered outcomes depending on severity and type. Beyond clinical features, several chemical properties of the administered drugs also contributed to classification. These included molecular weight, stereocenter counts, formal charge, and covalently-bonded unit count, which showed measurable impacts on outcome predictions. In particular, drugs with greater atom sterocenter count often shifted predictions toward fatal outcomes, reflecting potential links to pharmacokinetic or toxicological burden, whereas compounds with lower complexity leaned toward recovery. The influence of age and weight varied markedly by animal group. In companion animals, older age strongly increased the probability of Death predictions, while younger animals were more likely to recover. In livestock, body weight was the dominant demographic driver, with heavier animals trending toward fatal outcomes. Poultry, however, showed a more balanced effect, where both age and weight contributed modestly but consistently to predictions.

Across companion animals, livestock, and poultry, SHAP analysis highlights both protective and high-risk AEs influencing survival outcomes (Figure~\ref{fig:shap_summary_groups} (d-f)). Protective features consistently associated with recovery included stomach disorders, epidermal and dermal disorders, and injection/application site reactions, suggesting these events are less severe or more manageable. In contrast, strongly negative predictors of recovery (linked with death) varied across groups but showed common patterns: bronchial and lung disorders and general signs or symptoms were dominant in companion animals, while abnormal pathology reports (including bronchial, allergic, and abdominal conditions) were the most influential in livestock, and abnormal necropsy findings (intestinal, digestive, and vocalisation disorders) were critical in poultry. These results indicate that while mild digestive or dermal AEs correlate with survival, systemic pathology reports and respiratory or necropsy-related findings strongly predict fatal outcomes, emphasizing the need for early detection and management of severe systemic and respiratory complications across species.

Based on the Figure~\ref{fig:shap_summary_groups} (g-i), recovery in companion animals was most associated with active ingredients such as Grapiprant, Maropitant Citrate, and common parasiticides, while high risk of death was linked to Milbemycin A3 and anesthetic or antibiotic combinations. In livestock, model impact on Recovered cases was seen for active ingredients like Gamithromycin and Gentamicin Sulfate, whereas fatal outcomes were strongly tied to active ingredients such as Tilmicosin, Tildipirosin, and Terramycin, with Monensin showing both protective and harmful associations. For poultry, feed additives such as Roxarsone and ionophores like Lasalocid/Salinomycin aligned with recovery, but Monensin, Ivermectin, and Sulfadimethoxine were strong predictors of death, reflecting toxicity risks. Together, the figures highlight that while certain therapeutic and feed-related compounds promote recovery, others are consistently linked with fatal outcomes across species.

Overall, the SHAP analysis confirms that the model integrates clinical, demographic, and chemical information in ways that align with biological and pharmacological expectations: severe AEs, unfavorable age or weight profiles within each group, and high-complexity drug descriptors contribute to Death predictions, while milder conditions, favorable physiology, and simpler chemical properties correspond to recovery.

\section{Future Work}


This study focused on structured clinical, physicochemical properties, and demographic features in the OpenFDA CVM dataset and PubChem, but future work could be strengthened by adding protein-level and PK data. Protein descriptors—such as drug–target interactions, binding affinities, structural motifs, post-translational modifications, and protein–protein interaction networks (e.g., from STRING)- could reveal how drugs engage molecular pathways driving AEs. From a PK perspective, parameters such as absorption rate, bioavailability, distribution volume, plasma protein binding, half-life, clearance rate, and metabolic pathways could provide crucial insight into exposure–toxicity relationships \cite{riviere2017guide}. Integrating these protein-level and PK features with existing clinical AE data would transform the model from a purely statistical classifier into a mechanistically informed predictive framework. This could enable the detection of high-risk drug–event profiles even for drugs with limited historical AE reports by leveraging protein interaction networks and mechanistic analogies. Moreover, such integration would support explainable AI approaches, where feature importance could be tied back to protein targets and toxicokinetic principles, improving interpretability for veterinarians, regulators, and toxicologists.

\section{Conclusions}

This research demonstrates that integrating robust data preprocessing, advanced ML models, and explainable AI techniques can yield high-performing, clinically interpretable predictions of veterinary safety profiles and health outcomes. Using the FDA's OpenFDA veterinary PV data, we addressed real-world challenges including heterogeneous data formats, high-cardinality categorical variables, class imbalance, and incomplete records. Our results demonstrate that ensemble classifiers achieved the strongest overall performance, with CatBoost emerging as the best-performing and most consistently maintained model. When combined with resampling strategies and AUM-based pseudo-labeling, these methods further enhanced recall for fatal outcomes, reinforcing the framework’s utility for regulatory decision-making and food safety risk assessment. The inclusion of semi-supervised learning allowed the model to leverage unlabeled cases, increasing the diversity and representativeness of the training dataset without substantially degrading performance on recovered outcomes. Explainability tools, such as SHAP feature importance, revealed important clinical and toxicological patterns. They identified key predictors, including specific organ system disorders (e.g., bronchial and lung disorders, heart disorders) linked to fatal outcomes, animal demographic information, and notable drug characteristics. These findings not only confirm the accuracy of the model’s predictions but also highlight clinically relevant patterns that can support veterinarians, regulators, and researchers in understanding patterns across AEs, animal demographics, and drug-related attributes that are associated with reported outcomes. The proposed framework offers a scalable and transparent approach for integrating AI into veterinary PV workflows, enabling earlier detection of high-risk drug–event profiles. This capability can inform evidence-based regulatory actions, guide veterinary prescribing practices, and ultimately protect both animal welfare and human food safety.
\textbf{Code} will be released on GitHub, together with the full set of hyperparameters explored and the best configurations selected based on validation. 

\section*{Acknowledgments}

This work was supported by the USDA via the FARAD program (Award No.: 2022-41480-38135, 2023-41480-41034, 2024-41480-43679, and 2025-41480-45282) and its support for the 1DATA Consortium at Kansas State University.


This work also utilized resources available through the National Research Platform (NRP) at the University of California, San Diego. NRP has been developed and is supported in part by funding from the National Science Foundation, through awards 1730158, 1540112, 1541349, 1826967, 2112167, 2100237, and 2120019, as well as additional funding from community partners.
\bibliographystyle{IEEEtran}
\bibliography{references}

\end{document}